\documentclass{IEEEtran}

\usepackage{amsmath,amssymb,amsfonts}
\usepackage{algorithmic}
\usepackage{graphicx}
\usepackage{textcomp}
\usepackage[utf8]{inputenc}
\usepackage{algorithm2e}

\usepackage{graphicx}
\usepackage{subcaption}

\usepackage{xcolor}
\title{Reinforcement Learning Trajectory Generation and Control for Aggressive Perching on Vertical Walls with Quadrotors}

\author{Chen-Huan Pi, Kai-Chun Hu, Yu-Ting Huang, Stone Cheng}
\date{October 2019}

\begin{document}

\maketitle

\begin{abstract}
Micro aerial vehicles are widely being researched and employed due to their relative low operation costs and high flexibility in various applications. We study the under-actuated quadrotor perching problem, designing a trajectory planner and controller which generates feasible trajectories and drives quadrotors to desired state in state space. This paper proposes a trajectory generating and tracking method for quadrotor perching that takes the advantages of reinforcement learning controller and traditional controller. The trained low-level reinforcement learning controller would manipulate quadrotor toward the perching point in simulation environment. Once the simulated quadrotor has successfully perched, the relative trajectory information in simulation will be sent to tracking controller on real quadrotor and start the actual perching task. Generating feasible trajectories via the trained reinforcement learning controller requires less time, and the traditional trajectory tracking controller could easily be modified to control the quadrotor and mathematically analysis its stability and robustness. We show that this approach permits the control structure of trajectories and controllers enabling such aggressive maneuvers perching on vertical surfaces with high precision.
\end{abstract}

\begin{IEEEkeywords}
Aggressive flight, reinforcement learning, quadrotor
\end{IEEEkeywords}
\section{Introduction}
Micro Aircraft Vehicles (MAVs) are widely being researched and employed due to their relative low operation costs and high flexibility in various applications. The applications include search and rescue \cite{Choutri2020AFA}, wind turbine inspection \cite{Schafer2016MulticopterUA,Kaufman2018AutonomousQ3}, reconnaissance and mapping \cite{Durdevic2019LiDARAC}, target tracking, environmental monitoring, etc.

Most of the quadrotor is powered by battery for motor driving and generate lift, which gives a limited time for quadrotor tasks performing. Flying with a power cable connected to power supply on ground \cite{Fagiano2017SystemsOT,Bolognini2020LiDARBasedNO,Lee2015GeometricCF} can solve the problem but the quadrotor can only be operated in a specific and relative open area. If the quadrotor can perch on arbitrary angle surface such as vertical wall to reduce its battery consumption while keeping environment monitoring using on-board camera or even for battery charging. The mission time can be extended with a great amount of time.

The quadrotor is an under-actuated system, meaning it cannot control its position while maintaining arbitrary attitude. To successfully perch on the wall, quadrotors using gripper mechanism \cite{Thomas2017GraspingPA,Zhang2019CompliantBG, Kalantari2015AutonomousPA} to perch on vertical walls had been proposed. However, the perching device would cost additional energy consumption. Another approach for perching is using aggressive maneuver to land without adding mechanism on quadrotor.

There are a few other researches discussing perching of traditional quadrotors. To land a traditional quadrotor, the process can be separated into two different parts, 1) Trajectory planning to the target landing point. 2) Track the designed trajectory and land. To generate a feasible trajectory, Quadratic Program problem were formulated under the dynamics constrain of quadrotor \cite{Thomas2016AggressiveFW, Mellinger2012TrajectoryGA}.

\begin{figure}[t]
    \centering
    \includegraphics[width=\linewidth]{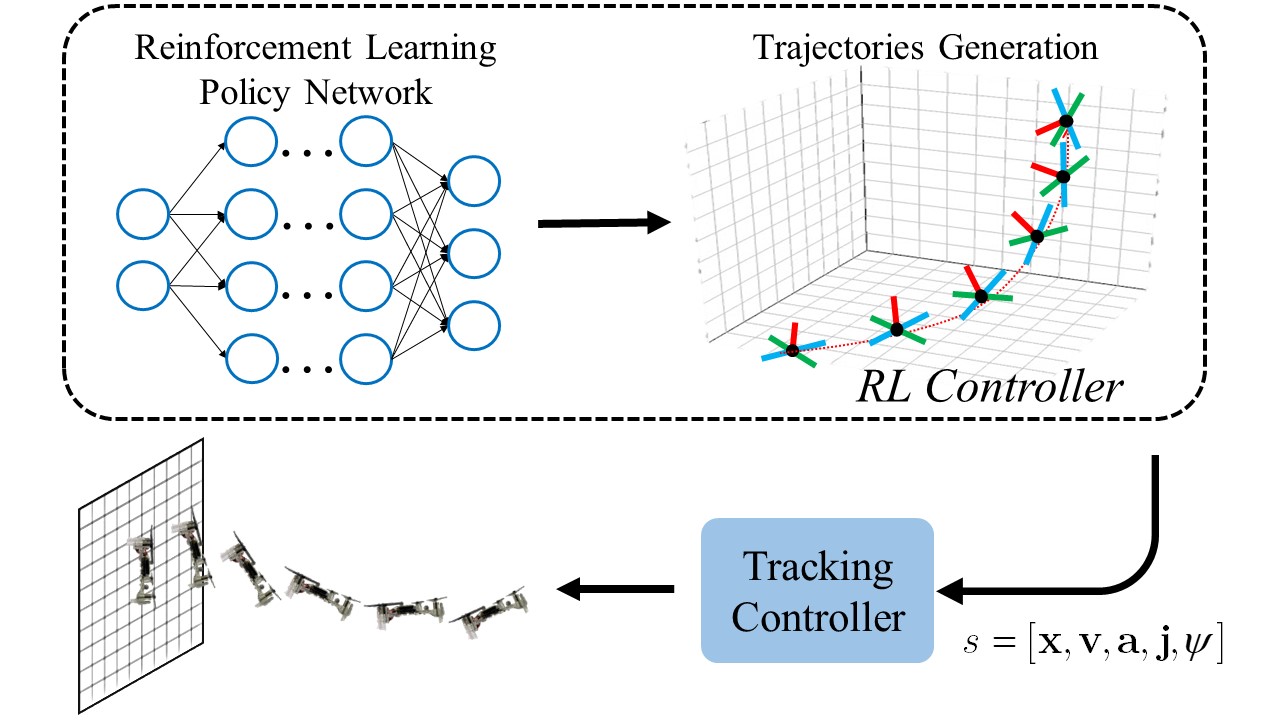}
    \caption{Aggressive perching maneuver using reinforcement learning trajectory planner and tracking controller}
    \label{fig:system_control_sturcture}
\end{figure}

Using Reinforcement Learning (RL) for low-level  quadrotor control has been proposed and shown its implementation result in real world quadrotor to stabilize and hover or fly on a specific trajectory \cite{Hwangbo2017ControlOA,Pi2020LowlevelAC}. The advantage of using RL controller is due to its nonlinearity and fast computation. However, there exist a gap showing on the tracking result while transferring the RL controller on quadrotors from simulation to real world experiment. In reality condition, the quadrotor hovering and tracking results didn't perform as same as the results in simulation. The reason is that the RL-based controller is sensitive to the deviation of model physical parameters such as take-off weight or motor characteristic. This situation can be harsher when considering aggressive maneuver using RL controlling the motors directly.

To fix this problem, we propose a method that takes the advantages of RL controller and traditional controller illustrated in Fig. \ref{fig:system_control_sturcture}. First, we train a RL controller to learn how to perch the quadrotor to the set point by directly command the thrust force of each motor. Second, we create a simulated quadrotor which has the same initial states as the real quadrotor. The trained low-level RL controller would manipulate quadrotor toward the landing point in simulation environment. Third, we extract the trajectory information in simulation and send it to traditional controller on real quadrotor. Generating feasible trajectories via the trained RL controller would require less time, and the traditional trajectory tracking controller could easily be modified to control the quadrotor and mathematically analysis its stability and robustness. Moreover, this combined structure of controller can be implement to different physical parameters or different type of vehicle.

\begin{figure}[t]
    \centering
    \includegraphics[width=\linewidth]{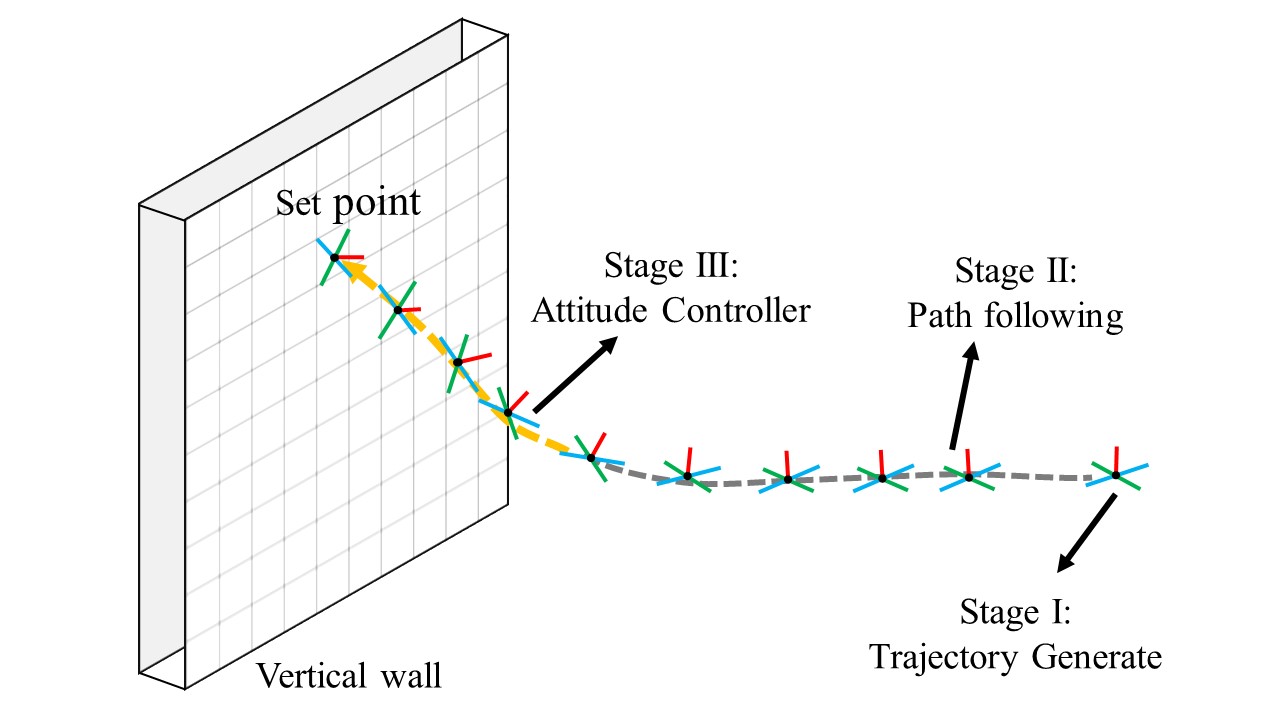}
    \caption{The quadrotor launch in random initial state. Stage I generate the trajectory of perching. In Stage II, the quadrotor start to track along the planned trajectory and switch to Stage III for attitude control while it approach to landing position.}
    \label{fig:Schematic_diagram}
\end{figure}
% we proposed a novel solution for quadrotor arbitrary landing task with a controller which takes the advantage of short computation time from RL and robust tracking from traditional controller. 

In Section \ref{sec:methodology}, the dynamics of quadrotor and reinforcement learning algorithm were presented. Section \ref{sec:Trajectory_Generating} and Section \ref{sec:controller} describes the RL perching training for trajectory generating and the traditional controller design to perform perching task. Section \ref{sec:experiment} shows the successful application of the combined structure to the quadrotor, and Section \ref{sec:conclusion} conclude the paper.

\section{Methodology}
\label{sec:methodology}
The first challenge of planning is that since the quadrotor is an under-actuated system, the designed trajectory needs to be feasible with respect to the system dynamic constrains. A possible technique is to solve a constrained optimization, for example:
\begin{equation}
    \label{equ:constrained optimization}
    \begin{split}
        \min_a & \int_{t_i}^{t_f}R(x, \dot x, a) dt + T(x(t_f))\\
        &\dot{x} = f(x,a)\\
        &x(t_i) = x_0
    \end{split}    
\end{equation}
where $R$ is run time cost, $T$ terminal cost, $f$ system dynamic, $a$ input action, and $x_0$ is the initial state. With this scheme, we can obtain a suitable trajectory $x(t)$ with respect to time.

% The next challenge is when tracking algorithm failed due to the system uncertainty\footnote{For example, a pulse disturbance will change the system state by unexpected reason.}, the failure recovery strategy also needs to be provided. Ideally, we can solve Problem (\ref{equ:constrained optimization}) by changing initial state $x_0$ to the current state, this technique know as model predictive control (MPC). Alternatively, we can simply give up current task and navigate to the designed initial state when current state is far from designed path. The first storage limited by computer power and the second storage far from elegance.

In this paper, we assuming dynamic trajectory planning problem can be described by a Markov decision processes (MDP), with this property the best decisions are independent to the history.
% >>>>>>>>>>>>>>>>>>>>> kai:edit
Next, we utilize reinforcement learning to search the feasible solutions in \eqref{equ:constrained optimization} while the searching results were memorized by an artificial neural network. With this scheme, the searching cost passes through the training phase.
% --------------------- original
% Next, we utilize the neural network capacity to store the proper decisions with respect to the corresponding states. It can be used as a control input or determine the next state and pass to the tracking algorithm. With this scheme, the searching cost are passing through the training phase.
% <<<<<<<<<<<<<<<<<<<<<

\subsection{Reinforcement Learning}
\label{subsec:reinforce learning}
Reinforcement learning (RL) is one of basic machine learning paradigms, aim to solve MDP through learning processes. In this section, we provide a short introduction to reinforcement learning.

A Markov decision processes is described by a 4-tuple $(\mathcal{S}, \mathcal{A}, \mathcal{P}, r)$ where $\mathcal{S}$ is the state space, $\mathcal{A}$ is action space, $\mathcal{P}=\mathcal{P}(s_{t+1}|s_t = s\in \mathcal{S}, a_t=a\in\mathcal{A})$ is transition probability and $r=r(s,a) \in \mathbb{R}$ is the reward function. The goal of MDPs is to find a policy 
\begin{equation}
    \label{equ:determinic policy}
    \pi=\pi(a\in \mathcal{A}|s)    
\end{equation}
that maximize cumulative discounted reward 
\begin{equation}
    \label{equ:cumulatived reward}
    \sum_{t=0}^\infty \gamma^t r(s_t, a_t),
\end{equation}
where $\gamma \in (0,1)$ is discounting factor.
To connect \eqref{equ:constrained optimization}, we write MDP statement as follow:
\begin{equation}
    \label{equ:MDPs}
    \begin{split}
        \max_{\pi} \mathbb{E}&\left[\sum_{t=0}^\infty \gamma^t r_t \Big| a_t \sim \pi(a|s_t) \right]\\
        & s_{t+1} \sim P(s| s_t, a_t)\\
        & s_0 \sim P_0(s)
    \end{split}
\end{equation}
where $r_t = r(s_t, a_t)$, $P_0$ is the initial state distribution.
For a given policy $\pi$,  the following recurrent relation hold
\begin{equation}
    \label{equ:recurrent value}
    V^\pi(s_t) = \mathbb{E}\left[r_t + \gamma V^\pi(s_{t+1}) \Big| a_t\sim \pi(a|s_t)\right],
\end{equation}
where
\begin{equation}
    \label{equ:value definition}
    V^\pi(s) = \mathbb{E}\left[\sum_{t=0}^\infty \gamma^t r_t \Big| a_t \sim \pi(a|s_t), s_0=s \right].
\end{equation}
The quantity 
\begin{equation}
    \label{equ:advantage}
    A^\pi(x,a) = \mathbb{E}\left[r_t+\gamma V^\pi(s_{t+1})| s_t = x, a_t = a\right] - V^\pi(s)
\end{equation}
can be used for determine that which action is good with respect to policy $\pi$. In fact, we can guarantee \cite{Pi2020LowlevelAC} that a policy $\pi$ is better than policy $\mu$ if
\begin{equation*}
    \forall (s,a) \in \mathcal{S}\times \mathcal{A}
\end{equation*}
\begin{subequations}
    \label{equ:dominate condition}
    \begin{equation}
        \left[\pi(a|s)-\mu(a|s)\right]A^\mu(s,a) > 0 
    \end{equation}
or
    \begin{equation}
        \left[\pi(a|s)-\mu(a|s)\right]A^\pi(s,a) > 0.
    \end{equation}
\end{subequations}
% >>>>>>>>>>>>> kai:edit
Therefore, for a policy search iteration, Equation \eqref{equ:dominate condition} provides an improvement guideline to distinguish whatever an action should be memorized or not.
% ------------- original
%Therefore, for a policy search iteration, Inequality \eqref{equ:dominate condition} provide a improvement guideline instate of evaluate $V^\pi$ on entire state domain.
%<<<<<<<<<<<<<<<

\subsection{Quadrotor Dynamics}
The simplest way to modeling a multi-rotor is considered as a rigid body dynamics
\begin{equation}
    \frac{d}{dt} \begin{pmatrix}
        \mathbf{R}\\
        \mathbf{\Omega}\\
        \mathbf{x}\\
        \mathbf{v} \\
        \mathbf{a}
    \end{pmatrix}
    =
    \begin{pmatrix}
        \mathbf{R}\hat{\mathbf{\Omega}}\\
        \mathbf{J}^{-1}\left(\tau-\mathbf{\Omega}\times\mathbf{J}\mathbf{\Omega} \right)\\
        \mathbf{v}\\
        \frac{1}{m} \mathbf{R}\left[0~ 0~T_z\right ]^T + \mathbf{g}\\
        \mathbf{s}
    \end{pmatrix}
    \label{eq:rotor_dynamic}
\end{equation}
% where the hat map ˆ· : R
% 3 → so(3) is defined by the condition
% that xyˆ = x × y for all x, y ∈ R
% 3
where $\mathbf{R}$ is rotation matrix, $\mathbf{\Omega}$ is angular velocity on body frame, and the \textit{hat map} $\hat{}:\mathbb{R}^3\rightarrow SO(3)$ is defined as the condition $\hat{x}y=x\times y,\forall x,y \in \mathbb{R}^3$ \cite{Lee2010GeometricTC}. The $\mathbf{x}$, $\mathbf{v}$, and $\mathbf{a}$ are position, velocity, acceleration with respect to inertia frame, $J$ is the momentum of inertia, $m$ is the mass of rigid body. $\mathbf{g}$ is the gravity acceleration. The force and moment model of each quadrotor motor can be simplified as
\begin{equation}
    F_i = k_F\omega^2_i, M_i = k_M\omega^2_i,
\end{equation}
where $\omega_i$ is the rotation speed of motors and $i=1,2,3,4$. $k_f$, $k_M$ are the coefficient of lift and moment.
The relation between moment $\tau$, force $T_z$ of quadrotor and motor speed $\omega_i$ can be written as
\begin{equation}
    \label{equ:thrust2generalforce matrix}
    \begin{pmatrix}
        \mathbf{\tau}\\
        T_z
    \end{pmatrix} =
    \begin{pmatrix}
        \frac{-l}{\sqrt{2}}k_F & \frac{l}{\sqrt{2}}k_F & \frac{l}{\sqrt{2}}k_F &\frac{-l}{\sqrt{2}}k_F\\
        \frac{l}{\sqrt{2}}k_F & \frac{-l}{\sqrt{2}}k_F & \frac{l}{\sqrt{2}}k_F &\frac{-l}{\sqrt{2}}k_F\\
        k_M & k_M & -k_M & -k_M \\
        -1&-1&-1&-1
    \end{pmatrix}\begin{pmatrix}
        \omega^2_1\\
        \omega^2_2\\
        \omega^2_3\\
        \omega^2_4
    \end{pmatrix}
\end{equation}
where $l$ is the physical dimension of quadrotor.

\subsection{Training and Inference}
Integrating \eqref{eq:rotor_dynamic}, \eqref{equ:thrust2generalforce matrix} and Euler method, we can obtain the discrete time transfer function $G$ of a quadrotor
\begin{equation}
    \label{equ:environment}
    s_{t+1} = G(s_t, a_t),
\end{equation}
where $s_t = (\mathbf{R},\mathbf{\Omega},\mathbf{x},\mathbf{v})_t$ and $a_t = (T_1, T_2, T_3, T_4)_t$.
We change our objective in terms of deterministic transition and the main objective become
\begin{equation}
    \label{equ:main objective}
    \begin{split}
        & \max_{\pi} V^\pi(s_0)\\
        & s_{t+1} = F(s_t, a_t)\\
        & s_0 \sim P_0(s)
    \end{split}
\end{equation}
in practice.

% \subsection{Training and Inference}
% >>>>>>>>>>>>>>>>>>> kai:edit
Actor-Critic architecture is used in this paper. The value network (Critic), was trained by a V-trace method \cite{Espeholt2018IMPALASD}, which is a modified temporal difference (TD) learning \cite{Sutton1988LearningTP}. The policy network (Actor), was trained according to \eqref{equ:dominate condition} through hinge loss. In addition, since \eqref{equ:dominate condition} can be satisfied by adjusting the likelihood per pair of state and action $(s,a)$, we replace the likelihoods by its log of likelihood for simplifying the complexity in gradient calculation. The summarized objective is listed as follows
\begin{equation}
    \begin{split}
        & \min_{\theta_\pi}\max\{0, \xi-\hat{A}_t\log\frac{\pi(a_t|s_t)}{\mu(a_t|s_t)}\},\\
        & \min_{\theta_v} \left|\hat{V}_t-V(s_t;\theta_v)\right|^2,\\
        & \hat{A}_t = r_t + \gamma V(s_{t+1}) - V(s_t) + \frac{\pi(a_{t+1}|s_{t+1})}{\mu(a_{t+1}|s_{t+1})} \hat{A}_{t+1},\\
        & \hat{V}_t = V(s_t; \theta_v) + \frac{\pi(a_{t}|s_{t})}{\mu(a_{t}|s_{t})}\hat{A}_t,\\
        & \mathcal{T} = \{(s_i, a_i, s_{i+1}, \mu(a_i|s_i))| i = 1\cdots T \} \in \mathcal{B},
    \end{split}
\end{equation}
where $\xi$ is the margin to make the samples that already satisfied \eqref{equ:dominate condition} not to provide losses for policy improvement \cite{Pi2020LowlevelAC}.

Notice that since the simplified rigid body dynamic is independent to the rotor number. The different physical configurations are only effect on $J$ and $m$, the trained RL policy could be implemented directly on to quadrotors.
Unfortunately, training RL controller on simplified ideal environment and transfer to real world still exists system uncertainty and system response latency. These factors may lead to the quadrotor become unstable especially under aggressive maneuver situation. To overcome this problem, we consider a hybrid control strategy and introduce a tracking algorithm in next section.

% \begin{algorithm}[]
% \label{alg:RL planner}
% \SetAlgoLined
% \KwResult{Happy Flight}
%  Given $N$ step ahead\;
%  \While{True}{
%     current state = state estimator\;
%     s = current state\;
%     \While{$i < N$}{
%         $a = \pi(s)$\;
%         $s = F(s,a)$\;
%     }
%     goal state = s\;
%     tracker(goal state, current state)\;
%  }
%  \caption{Happy Algorithm}
% \end{algorithm}

\section{Simulation Environment and Trajectory Generating Result}
\label{sec:Trajectory_Generating}

In this section, we used the aforementioned dynamic equations and reinforcement algorithms to train a quadrotor perching policy controlling the four motors output directly which can successfully landing on the vertical wall. The reward function for RL training is designed as follows 
\begin{equation}
\begin{aligned}
    \text{reward}=-\begin{bmatrix}
                w_1e^{-d_x} & w_2 & w_3
                \end{bmatrix}
                \begin{bmatrix}
            \parallel \mathbf{R}_e\parallel & \parallel \mathbf{x}_e\parallel & \parallel a\parallel
                \end{bmatrix}^T,
\end{aligned}
\end{equation}
where $\mathbf{R}_e$, $\mathbf{x}_e$, and $a$ are angle error, position error, and action output. $w_1$ to $w_3$ are the weights of the error, and $d_x$ is the distance between quadrotor to the perching point on $x$-axis.
For RL value and policy neural network construction, 3-layer with 32 and 128 nodes were used. The policy network outputs four motors thrust command to control the quadrotor perching task. 

% present our simulation environment for trajectory learning and generating. We constructed a quadrotor simulator written in Python based on the dynamic model outlined in Sec. \ref{sec:methodology} which use the simplest dynamic model. The simulated environment only takes the effects of gravity and the forces generated by the motors. 

Once the quadrotor lands successfully in the simulator controlled by RL policy neural network, we extract the trained controller for generating feasible trajectory. The trajectory contains position, velocity, acceleration, jerk, and yaw angle information and use designed traditional controller to track along the planned path.

\begin{figure}
    \centering
    \includegraphics[width=0.48\textwidth,trim={1.5cm 3cm 1.5cm 3cm},clip]{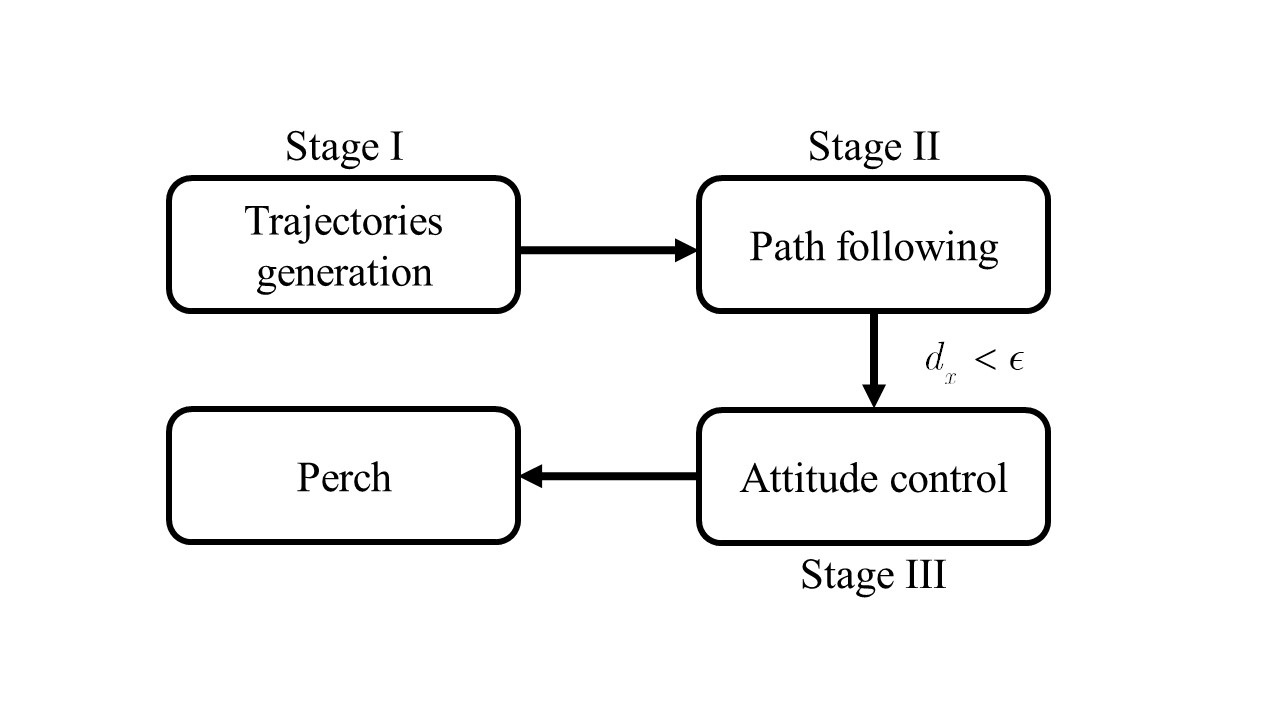}
    \caption{Three stages of perching control sequence}
    \label{fig:sequence}
\end{figure}

We design a sequence for controlling the quadrotor during operating the mission and can be separated into following three stages and shown in Fig.\ref{fig:sequence}:

\begin{itemize}
\item Stage I: Trajectories generation
\item Stage II: Path following
\item Stage III: Attitude control
\end{itemize}

In stage I, feasible trajectory would be generated from the current state of quadrotor and can successfully perch to desired point eventually. During the path following stage, the controller keeps tracking on the desired path with the information of position, velocity, acceleration, jerk, and yaw angle until the quadrotor approaches the launch point. As the distance $d_x$ is shorter than designed threshold $\epsilon$, quadrotor switch to stage III and attitude control would be triggered. During the attitude control stage, the controller attempt to stable the quadrotor to final pose and perch on the set point of a vertical plane. While the quadrotor reaching the original point with pitch angle of $90^{\circ}$, it would be considered to be a successful cases to landing onto the vertical surface.

\section{Controller Design}
\label{sec:controller}
In this section, we introduce our controller for trajectory tracking along perching neural network behavior.
The tracking and perching consist of two controllers as follows,

1. Trajectory Tracking Control: controlling the quadrotor center of mass to follow the three-dimensional path including position, velocity, acceleration, and jerk information from RL trajectory generator.

2. Attitude Control: controlling the quadrotor to desired roll, pitch and yaw angle from trajectory tracking controller command.

\subsection{Trajectory Tracking Control}
The tracking controller is based on proportional-derivative (PD) controller and can be written as following expression,
\begin{equation}
    \mathbf{a}_c = \mathbf{K_x} (\mathbf{x}_{ref}-\mathbf{x}_f) + \mathbf{K_v} (\mathbf{v}_{ref}-\mathbf{v}_f) + \mathbf{a}_{ref}
\end{equation}
where $\mathbf{a_c}$ is the desired acceleration vector of quadrotor, $\mathbf{x}_{ref}, \mathbf{v}_{ref},\mathbf{a}_{ref}$ are the control references from trajectory generator, $\mathbf{x}_f$, and $\mathbf{v}_f$ are the feedback, and $\mathbf{K_x},\mathbf{K_v}$ are the gains of each component.
The total thrust $T_c$ and desired attitude expressed in rotation matrix form $(\mathbf{R}_c)_{Traj}$ can be derived as
\begin{equation}
\begin{aligned}
    T_c &= -m\left \| \mathbf{a}_c \right \| \cdot \mathbf{b_z}\\
    (\mathbf{R}_c)_{Traj} &= \left [ \mathbf{b_{x}}_c~ \mathbf{b_{y}}_c~ \mathbf{b_{z}}_c \right ],
\end{aligned}
\end{equation}
where
\begin{equation}
\begin{aligned}
    \mathbf{e}_y &= [-\sin\psi_{ref}, \cos\psi_{ref}, 0]^T\\
    \mathbf{b_{z}}_c &= \frac{\mathbf{a}_c}{\|\mathbf{a}_c\|}\\
    \mathbf{b_{x}}_c &= \frac{\mathbf{e}_y \times\mathbf{b_{z}}_c}{\|\mathbf{e}_y \times \mathbf{b_{z}}_c\|}\\
    \mathbf{b_{y}}_c &= \frac{\mathbf{b_{z}}_c \times\mathbf{b_{x}}_c}{\|\mathbf{b_{z}}_c \times \mathbf{b_{x}}_c\|}
\end{aligned}
\end{equation}
$\psi_{ref}$ is the desired heading angle, and $\|\mathbf{e}_y \times \mathbf{b_{z}}_c\|>0$ is always greater than 0 because the quadrotor would not operate upside down in trajectory tracking control.

\subsection{Attitude Control}
The attitude controller not only receive the $(\mathbf{R}_c)_{Traj}$ command from trajectory tracking control but also switch to final perching attitude $(\mathbf{R}_c)_{P}$. The controller consists of dual-loop control structure. The outer loop P control in \eqref{eq:outer_loop} determines the desired angular velocity of quadrotor, where $\mathbf{R}_c$ switches according to the distance of quadrotor to perching point on $x$-axis $d_x$.
\begin{equation}
    \mathbf{R}_c=\left\{\begin{matrix} (\mathbf{R}_c)_{Traj} & ,d_x\geq\epsilon \\ (\mathbf{R}_c)_{P} & ,0<d_x<\epsilon \end{matrix}\right.
    \label{eq:switching_condition}
\end{equation}
The error of rotational angle between $\mathbf{R}_c$ and measurement feedback $\mathbf{R}_f$ is given by
\begin{equation}
\begin{aligned}
    \mathbf{e_R} &= -0.5(\mathbf{R}_c^T\mathbf{R}_f-\mathbf{R}_f^T\mathbf{R}_c)^\vee
\end{aligned}
\end{equation}
where the \textit{vee map} $\vee$ is the mapping $\mathrm{SO}(3)\rightarrow \mathbb{R}^3$ \cite{Lee2010GeometricTC}, and the desired angular velocity $\mathbf{\Omega}_c$ is derived as
\begin{equation}
\begin{aligned}
    \mathbf{\Omega}_c &= \mathbf{K_R} \mathbf{e_R}+\mathbf{\Omega}_{ref}.
    \label{eq:outer_loop}
\end{aligned}
\end{equation}
The $\mathbf{\Omega}_{ref}$ is the feed-forward term given from trajectory tracking controller, which use the jerk of the trajectory to compensate angular velocity command of the quadrotor \cite{Tal2018AccurateTO}.

The inner loop uses a PID controller for desired angular acceleration shown in \eqref{eq:inner_loop}.
\begin{equation}
\begin{aligned}
    \mathbf{e_\Omega} &= -\mathbf{\Omega}_f + \mathbf{R}_f^T \mathbf{R}_c\mathbf{\Omega}_c\\
    \dot{\mathbf{\Omega}}_c &= \mathbf{K_p}\mathbf{e_\Omega} + \mathbf{K_i} \int\mathbf{e_\Omega} + \mathbf{K_d}\mathbf{\dot{e}_\Omega}
    \label{eq:inner_loop}
\end{aligned}
\end{equation}
where $\mathbf{\Omega}_f$ is the feedback of quadrotor angular velocity.

The command of rotor speed is determined according to \eqref{eq:rotor_dynamic} \eqref{equ:thrust2generalforce matrix} with $T_c$ and $\dot{\mathbf{\Omega}}_c$.

\section{Simulation Results}
\label{sec:experiment}
\begin{figure}
    \centering
    \begin{subfigure}[b]{0.49\linewidth}
        \includegraphics[width=\linewidth]{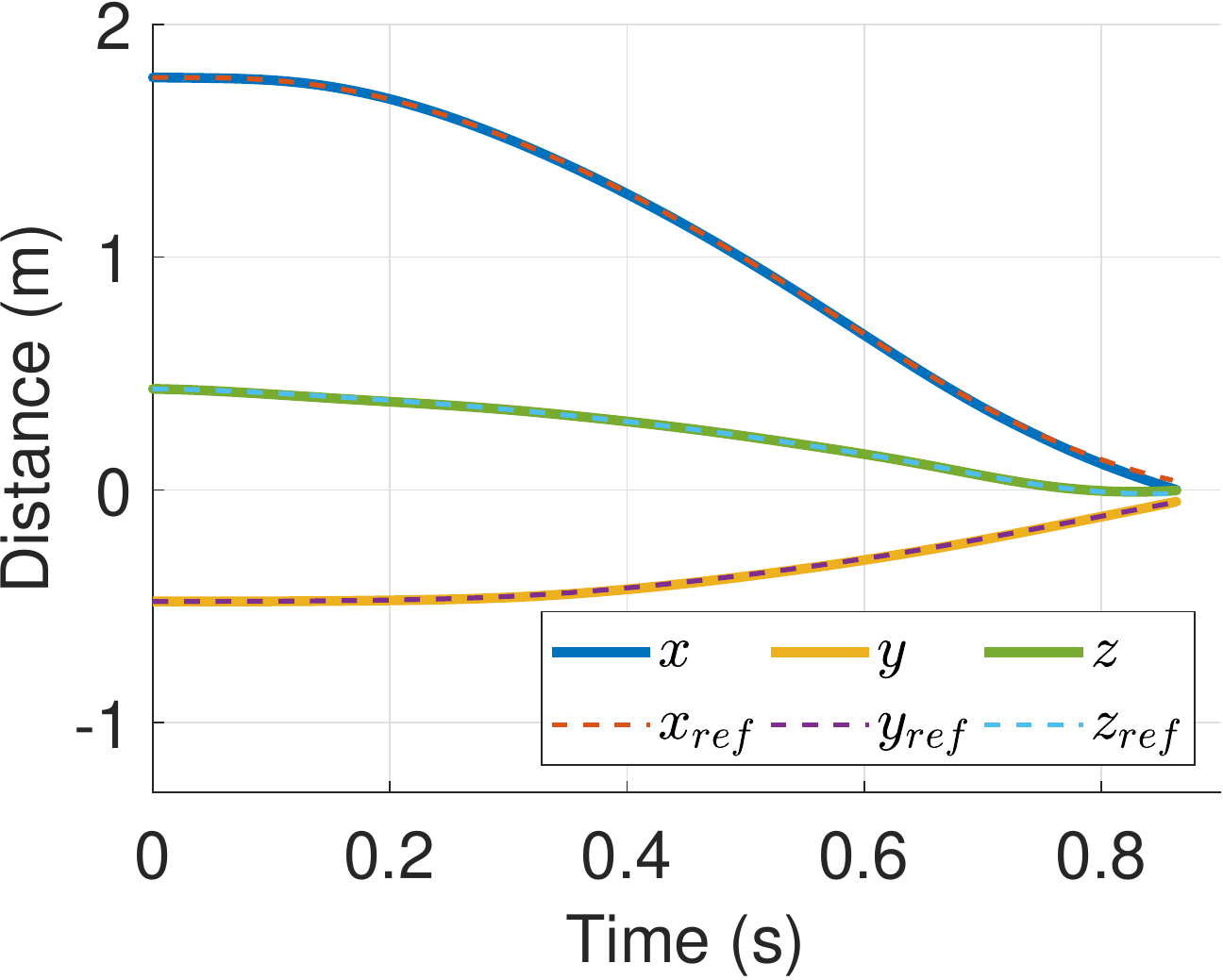}
        % \caption{Roll and pitch tracking in simulation.}
        \caption{Case 1}
    \end{subfigure}
    \begin{subfigure}[b]{0.49\linewidth}
        \includegraphics[width=\linewidth]{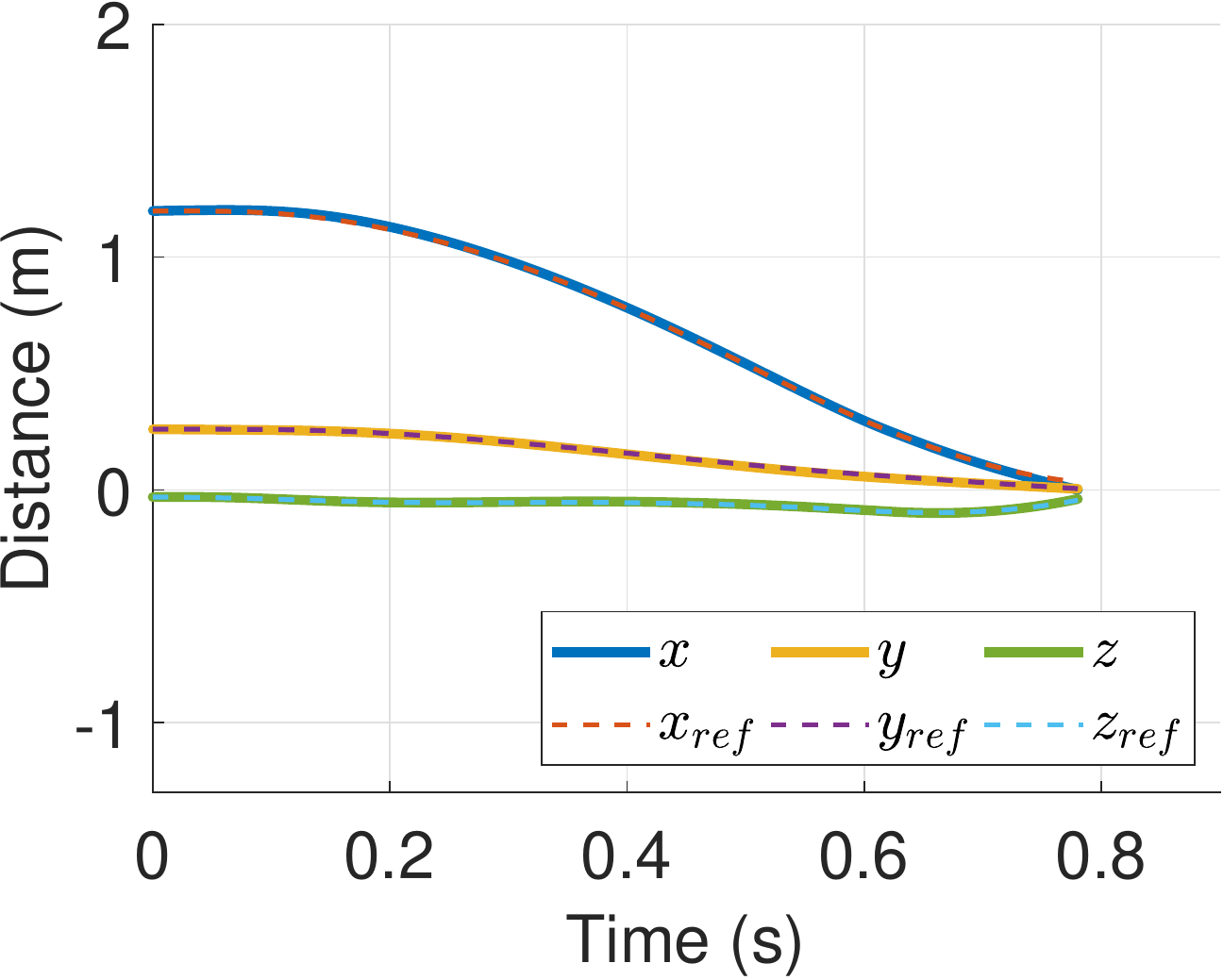}        % \caption{Roll and pitch tracking in experiment.}
        \caption{Case 2}
    \end{subfigure}
    
    \begin{subfigure}[b]{0.48\linewidth}
        \includegraphics[width=\linewidth]{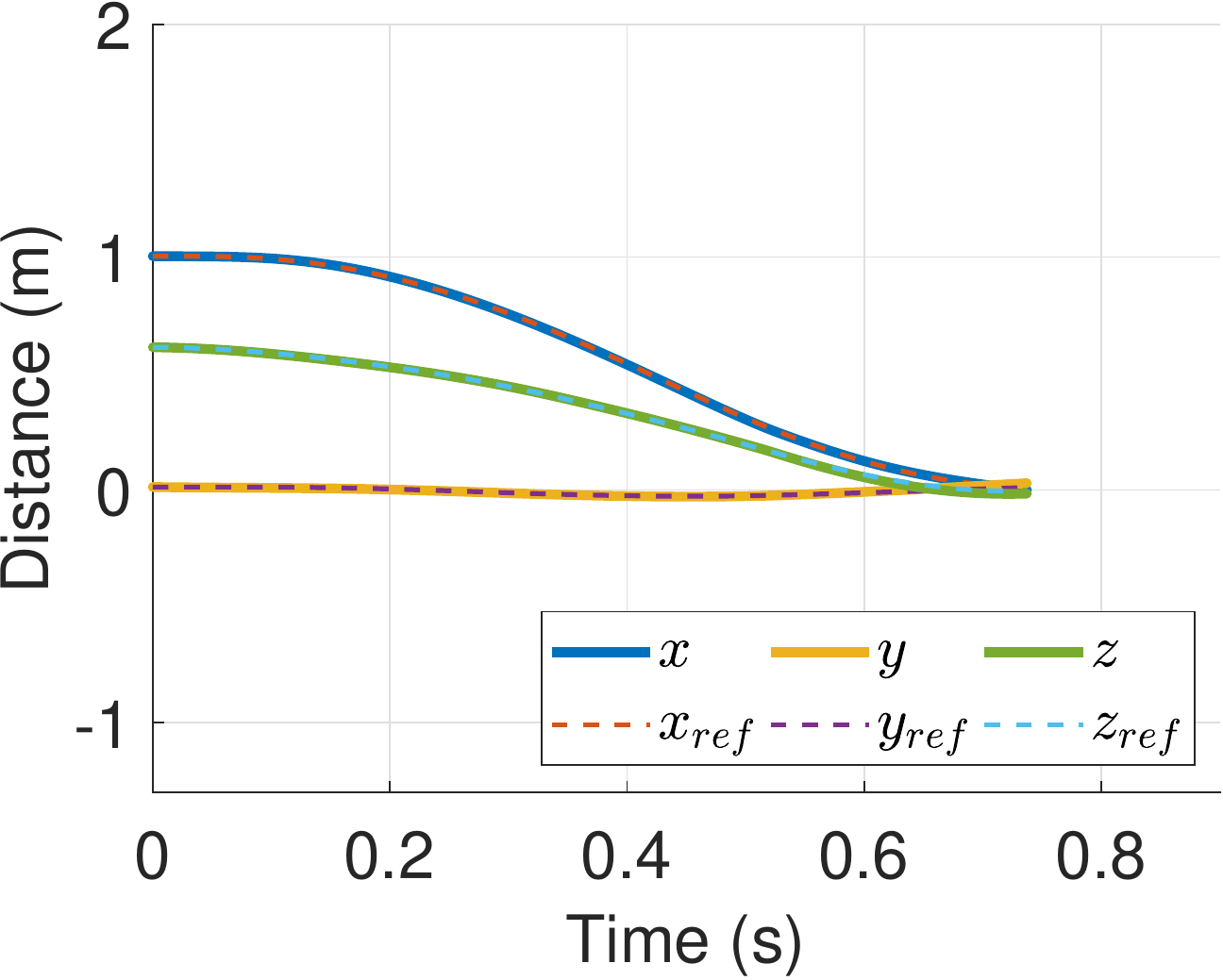}        % \caption{ in simulation.}
        \caption{Case 3}
    \end{subfigure}
    \begin{subfigure}[b]{0.48\linewidth}
        \includegraphics[width=\linewidth]{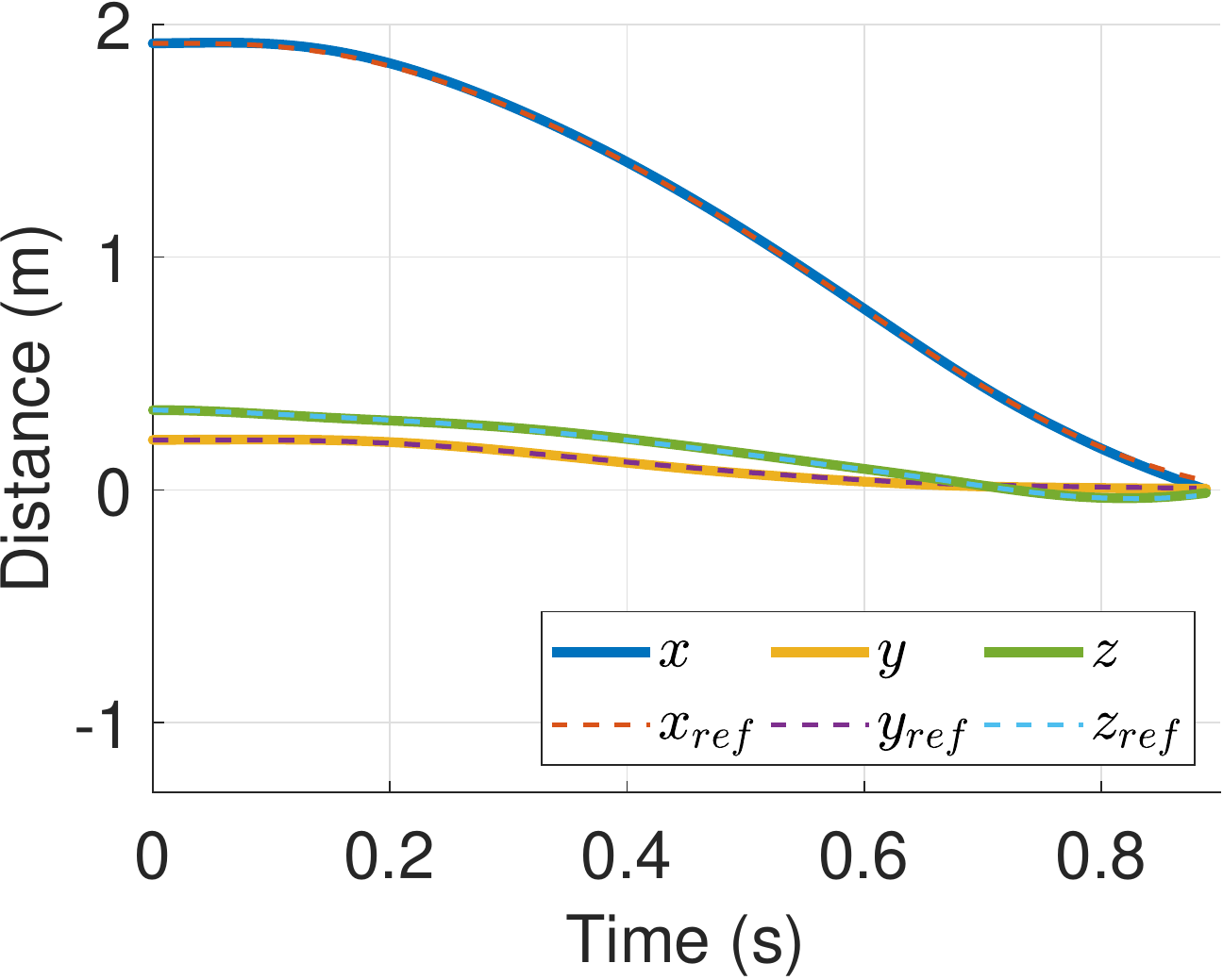}        % \caption{Position tracking in experiment.}
        \caption{Case 4}
    \end{subfigure}
    
    \caption{Four cases of flight result start from different initial condition.}
    \label{fig:multi_sim}
\end{figure}

To verify that our proposed perching trajectory generating and tracking structure could successfully landing on the wall, we conduct 50 flight tests starting from arbitrary initial position and velocity in a $1\mathrm{m} \times 1\mathrm{m} \times 1\mathrm{m}$ space in front of the perching point in the simulation environment illustrates in Fig. \ref{fig:Schematic_diagram}. All trails perch to the set position successfully. Fig. \ref{fig:multi_sim} shows four flight results start from different initial condition. Table. \ref{tab:landing_result} shows the mean and standard deviation (SD) of the distance error and pitch angle while the The vehicles touch the wall. The simulation experiment shows while the quadrotor reach to $x=0$, the mean distance error to the landing point on $y$ and $z$ is $-0.47$ cm and $-1.74$ cm. The mean pitch angle is $-1.17^{\circ}$.
\begin{table}
\centering
\caption[caption]{Simulation results of sampling 50 trails and the mean contact position and SD on $y,z$-axis and pitch angle $\theta$ while the quadrotor reach position $x=0$.}
\label{tab:landing_result}
\begin{tabular}{|c|c|c|}
\hline
            & Mean          & SD           \\ \hline
$y$-axis    & $-0.47$ cm      & 0.2 cm       \\ \hline
$z$-axis    & $-1.74$ cm      & 0.21 cm      \\ \hline
pitch angle & $88.83^{\circ}$ & $0.62^{\circ}$ \\ \hline
\end{tabular}
\end{table}

\section{Conclusion}
\label{sec:conclusion}

In this paper, we present a quadrotor perching control structure which combining RL trajectory planning and traditional tracking controller. The complex nonlinear dynamic constrain optimization problem can be solved by using trained RL trajectory planner to find a feasible path rapidly. A well designed traditional controller can guarantee the stability of quadrotor. The control structure can make up the uncertainty using RL control policy output directly to motors when it has disturbance in operation or modeling imperfection in RL training process. The perching simulation in Section \ref{sec:experiment} demonstrates the quadrotor follows the trajectory generated by RL control policy and perches to designed point with standard deviation $0.2$ cm on $y,z$-axis and pitch angle $88.83^{\circ}$ in average on the wall.

Future work will focus on the real-world implementation, verify the overall control structure perching performance, and observing the robustness while facing disturbance and model uncertainties. However, the long term goal is to perform tasks physical contact such as battery charging or environment monitoring.
\bibliographystyle{plain}
\bibliography{main}

\begin{thebibliography}{10}

\bibitem{Bolognini2020LiDARBasedNO}
Michele Bolognini and L.~Fagiano.
\newblock Lidar-based navigation of tethered drone formations in an unknown
  environment.
\newblock {\em ArXiv}, abs/2003.12981, 2020.

\bibitem{Choutri2020AFA}
K.~Choutri, M.~Lagha, and L.~Dala.
\newblock A fully autonomous search and rescue system using quadrotor uav.
\newblock 2020.

\bibitem{Durdevic2019LiDARAC}
P.~Durdevic, D.~Ortiz-Arroyo, and Z.~Yang.
\newblock Lidar assisted camera inspection of wind turbines: Experimental
  study.
\newblock {\em 2019 1st International Conference on Electrical, Control and
  Instrumentation Engineering (ICECIE)}, pages 1--7, 2019.

\bibitem{Espeholt2018IMPALASD}
Lasse Espeholt, Hubert Soyer, R.~Munos, K.~Simonyan, V.~Mnih, Tom Ward, Yotam
  Doron, Vlad Firoiu, T.~Harley, Iain Dunning, S.~Legg, and K.~Kavukcuoglu.
\newblock Impala: Scalable distributed deep-rl with importance weighted
  actor-learner architectures.
\newblock {\em ArXiv}, abs/1802.01561, 2018.

\bibitem{Fagiano2017SystemsOT}
L.~Fagiano.
\newblock Systems of tethered multicopters: Modeling and control design.
\newblock {\em IFAC-PapersOnLine}, 50:4610--4615, 2017.

\bibitem{Hwangbo2017ControlOA}
J.~Hwangbo, I.~Sa, R.~Siegwart, and M.~Hutter.
\newblock Control of a quadrotor with reinforcement learning.
\newblock {\em IEEE Robotics and Automation Letters}, 2:2096--2103, 2017.

\bibitem{Kalantari2015AutonomousPA}
A.~Kalantari, Karan Mahajan, D.~Ruffatto, and M.~Spenko.
\newblock Autonomous perching and take-off on vertical walls for a quadrotor
  micro air vehicle.
\newblock {\em 2015 IEEE International Conference on Robotics and Automation
  (ICRA)}, pages 4669--4674, 2015.

\bibitem{Kaufman2018AutonomousQ3}
Evan Kaufman, Kuya Takami, Zhuming Ai, and Taeyoung Lee.
\newblock Autonomous quadrotor 3d mapping and exploration using exact occupancy
  probabilities.
\newblock {\em 2018 Second IEEE International Conference on Robotic Computing
  (IRC)}, pages 49--55, 2018.

\bibitem{Lee2015GeometricCF}
Taeyoung Lee.
\newblock Geometric controls for a tethered quadrotor uav.
\newblock {\em 2015 54th IEEE Conference on Decision and Control (CDC)}, pages
  2749--2754, 2015.

\bibitem{Lee2010GeometricTC}
Taeyoung Lee, M.~Leok, and N.~McClamroch.
\newblock Geometric tracking control of a quadrotor uav on se(3).
\newblock {\em 49th IEEE Conference on Decision and Control (CDC)}, pages
  5420--5425, 2010.

\bibitem{Mellinger2012TrajectoryGA}
Daniel Mellinger, Nathan Michael, and Vijay Kumar.
\newblock Trajectory generation and control for precise aggressive maneuvers
  with quadrotors.
\newblock {\em The International Journal of Robotics Research}, 31:664 -- 674,
  2012.

\bibitem{Pi2020LowlevelAC}
Chen-Huan Pi, Kai-Chun Hu, Stone Cheng, and I-Chen Wu.
\newblock Low-level autonomous control and tracking of quadrotor using
  reinforcement learning.
\newblock {\em Control Engineering Practice}, 95:104222, 2020.

\bibitem{Schafer2016MulticopterUA}
Bjorn~E. Schafer, D.~Picchi, T.~Engelhardt, and D.~Abel.
\newblock Multicopter unmanned aerial vehicle for automated inspection of wind
  turbines.
\newblock {\em 2016 24th Mediterranean Conference on Control and Automation
  (MED)}, pages 244--249, 2016.

\bibitem{Sutton1988LearningTP}
R.~Sutton.
\newblock Learning to predict by the methods of temporal difference learning.
\newblock 1988.

\bibitem{Tal2018AccurateTO}
Ezra Tal and Sertac Karaman.
\newblock Accurate tracking of aggressive quadrotor trajectories using
  incremental nonlinear dynamic inversion and differential flatness.
\newblock {\em 2018 IEEE Conference on Decision and Control (CDC)}, pages
  4282--4288, 2018.

\bibitem{Thomas2017GraspingPA}
Justin Thomas.
\newblock Grasping, perching, and visual servoing for micro aerial vehicles.
\newblock 2017.

\bibitem{Thomas2016AggressiveFW}
Justin Thomas, M.~Pope, Giuseppe Loianno, E.~Hawkes, M.~A. Estrada, Hao Jiang,
  M.~Cutkosky, and V.~Kumar.
\newblock Aggressive flight with quadrotors for perching on inclined surfaces.
\newblock {\em Journal of Mechanisms and Robotics}, 8:051007, 2016.

\bibitem{Zhang2019CompliantBG}
Haijie Zhang, Jiefeng Sun, and Jianguo Zhao.
\newblock Compliant bistable gripper for aerial perching and grasping.
\newblock {\em 2019 International Conference on Robotics and Automation
  (ICRA)}, pages 1248--1253, 2019.

\end{thebibliography}
\end{document}